\documentclass{article}
\usepackage[utf8]{inputenc}
\usepackage{graphicx}
\usepackage{amsmath}
\usepackage{amssymb}
\usepackage{amsfonts}
\usepackage[hidelinks]{hyperref}

\title{Leveraging Systematic Knowledge about 2D Transformations}
\author{Jiachen Kang$^1$ \and Wenjing Jia$^1$ \and Xiangjian He$^1$}
\date{$^1$School of Electrical and Data Engineering, University of Technology Sydney}

\begin{document}

\maketitle

\begin{abstract}
    The existing deep learning models suffer from out-of-distribution (\textit{o.o.d.}) performance drop in computer vision tasks.
    In comparison, humans have a remarkable ability to interpret images, even if the scenes in the images are rare, thanks to the systematicity of acquired knowledge.
    This work focuses on 1) the acquisition of systematic knowledge about 2D transformations, and 2) architectural components that can leverage the learned knowledge in image classification tasks.
    With a new training methodology based on synthetic datasets that are constructed under a causal framework, the deep neural networks acquire knowledge from semantically different domains (\textit{e.g.} even from noise), and exhibit certain level of systematicity in parameter estimation experiments.
    Based on this, a novel architecture is devised consisting of a \textsc{classifier}, an \textsc{estimator} and an \textsc{identifier} (abbreviated as ``CED''). 
    By emulating the ``hypothesis-verification'' process in human visual perception, CED improves the classification accuracy significantly on test sets under covariate shift.
\end{abstract}

\section{Introduction}\label{s:intro}

Machine learning algorithms based on deep neural networks (DNNs) have made dramatic progress in the field of computer vision in the last decade.
Most of these algorithms strongly rely on the assumption of \textit{i.i.d.}, \textit{i.e.}, the training data and test data are independent and identically distributed.
In practice, however, the \textit{i.i.d.}~assumption can be easily violated due to covariate shift in test datasets~\cite{alcorn2019strike, barbu2019objectnet, goodfellow2014explaining, jo2017measuring}, which causes significant performance drop of the models learned from the training set.
This is investigated as the \textit{o.o.d.}~generalization problem, which has become one of the main challenges that the deep learning community encounters nowadays. 
One of the common stopgaps for this problem is to continuously expand the size of datasets, in order to strengthen the learned invariance of the target objects, by getting rid of other mechanisms or factors of variation.
For example, ImageNet~\cite{deng2009imagenet}, which is a typical dataset for training classification and detection algorithms, contains more than 14 million images.
Even so, popular classification models trained with ImageNet have experienced $40-45\%$ performance drop when tested on ObjectNet, a bias-controlled dataset~\cite{barbu2019objectnet} that produces thousands of images with 600 combinations of parameters, by intervening only on three mechanisms in the photo generation process. 
This implies that if we try to construct a big enough dataset to approximate the distribution of real-world data, by considering all possible combinations of parameters of mechanisms, the number of required data points would be nearly infinite.

\begin{figure}[bt] 
    \vskip 0.1in
    \begin{center}
    \centerline{\includegraphics[width=0.6\columnwidth]{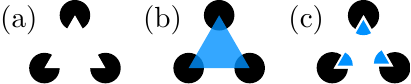}} 
    \caption{What is in image (a)? There are at least two ways to interpret it, \textit{i.e.}, (b) three black circles partly \emph{covered} by a white triangle, or (c) three black circles with a \emph{notch} on each of them. (The former one may have a stronger tendency in perception, according to the Gestalt principles~\cite{koffka2013principles}.)}
    \label{i:f1_gestalt}
    \end{center}
    \vskip -0.2in
\end{figure}

Human beings, in comparison, have powerful \textit{o.o.d.}~generalization abilities that enable us to recognize objects based on efficient learning. 
Extensive studies have shown that learned knowledge can be flexibly reused by infants in novel scenarios~\cite{teglas2011pure, battaglia2013simulation, lake2017building, scholkopf2021toward}. 
This is analogous to algebraic operations~\cite{marcus2003algebraic} where symbolic variables are manipulated in computational processes.
This can be a crucial explanation for the generalization ability.
To illustrate this, if we look at Fig.~\ref{i:f1_gestalt}(a)~\cite{lehar2003world}, at least two interpretations can be made (Fig.~\ref{i:f1_gestalt}(b) and \ref{i:f1_gestalt}(c)), based on the same observation. 
This simple example illustrates a typical process of image perception, in which causal inference (in the anti-causal direction) is made by utilizing the mechanisms of either occlusion or notching on variables of circles and/or triangles.
Specifically, the process consists of a hypothesis (of the content of three circles and a triangle) and the verification (whether a figure like this can be generated by \emph{covering} the triangle over the circles).
If another hypothesis (\textit{e.g.} of just three circles) and a corresponding verification (by \emph{making a notch} in each of them) can be made, the figure still makes sense to us.
This ``hypothesis-verification'' process in human visual perception is discussed in detail in~\cite{marcel1983conscious}.
It can be noticed that mechanisms in the image generation process (\textit{e.g.}~occlusion or notching) are crucial in human visual perception. 
How an image is perceived relies on our knowledge of various mechanisms, rather than knowledge of images that are previously seen (which is the way that existing machines operate).
It can also be noticed that our knowledge about occlusion or notching is universal and independent of the domain of variables.
This generalization is also referred to as systematicity~\cite{fodor1988connectionism}.
Based on the above analysis, it can be inferred that it is the systematicity of acquired knowledge that enables human beings to take mechanisms into consideration in visual perception, and thus achieve excellent \textit{o.o.d.}~generalization ability.

While children have plenty of time to gain systematic knowledge and physical mechanisms through observations and experiments~\cite{stahl2015observing, schulz2007preschool, cook2011science}, which build foundations for object perception and future knowledge acquisition~\cite{schmidt1986development, spelke1990principles, lake2017building}, existing machine learning models rarely have opportunities to do so.
One of the main reasons is that current datasets for visual learning inevitably introduce confounding mechanisms, which makes it difficult for models to learn unbiased representations and acquire systematic knowledge.
Additionally, most of the studies focus on learning the invariance of objects of interest, and neglect the fact that other mechanisms also provide necessary information for perception, as shown in the previous example.

This work does not only focus on learning the invariance of objects of interest, we also pay attention to other mechanisms.
Therefore, empirical studies are conducted to learn knowledge about mechanisms of 2D transformations (such as rotation, scaling and translation) using DNNs, in order to answer the following questions: \\
\begin{enumerate}
\item Whether the knowledge of these mechanisms learned by machines can exhibit some level of systematicity? If so,
\item whether the knowledge can be leveraged to facilitate image classification tasks like humans? 
\end{enumerate}


In order to answer the first question, it should be made clear what we mean by \emph{the knowledge of a mechanism}.
As human beings, for example, if we have learned the knowledge of 2D rotation, it means that for any image, (with a proper tool), (a) we can rotate the image at will, and (b) we are able to determine whether (and even how many degrees) the image has been rotated.
Obviously, the knowledge we know about 2D rotation generalizes systematically and is independent of the domain of images.
For transformations studied in this work, the affine transformation functions are in accord with the description in (a), and are used in the architecture as a tool to make precise operations\footnote{It does not imply that transformation operations cannot be learned from data. 
Generative models which is beyond the scope of this study, have been studied in various tasks~\cite{mirza2014conditional, vincent2010stacked}.}.
Therefore, our main purpose is the learning of the latter aspect (b).
To achieve this, synthetic datasets are designed under a causal framework as the training datasets. 
Specifically, the datasets are composed of pairs of images, which are before and after the transformations, respectively.
It has been found that with this training methodology, the transformation parameters can be estimated more accurately and stably even on \textit{o.o.d.} data that are semantically different.

\begin{figure*}[t] 
    \vskip 0.1in
    \begin{center}
    \centerline{\includegraphics[width=1\textwidth]{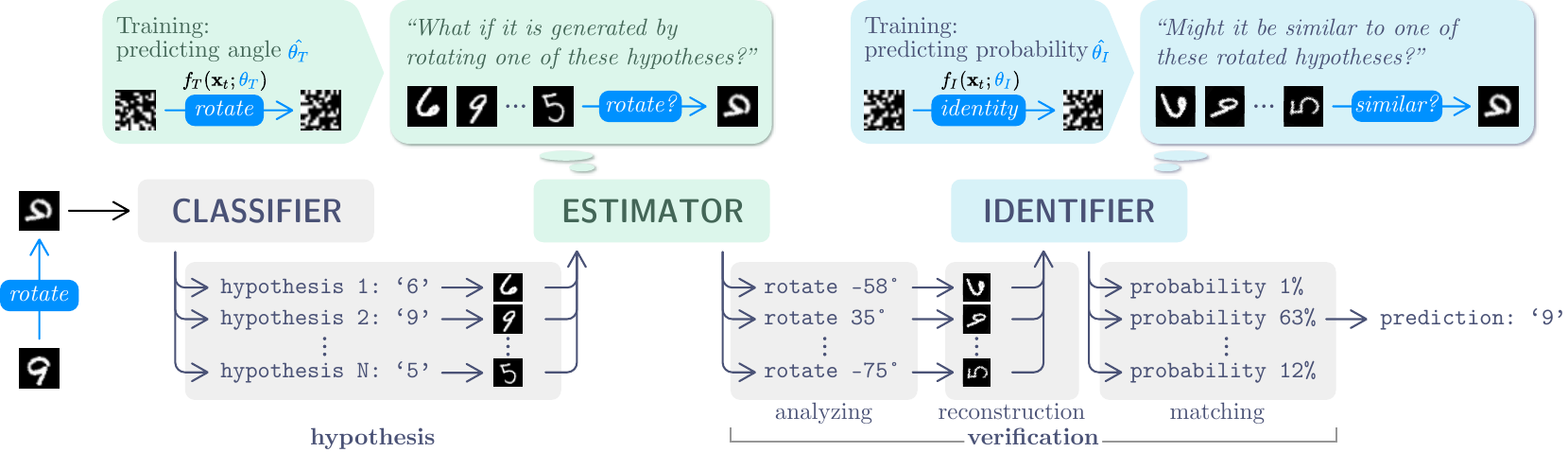}} 
    \caption{The CED architecture. Potential classes are hypothesized by the \textsc{classifier} $C$, and verification on these classes is made by the \textsc{estimator} $E$ and the \textsc{identifier} $D$ through the pipeline of (1) analyzing possible transformations, (2) reconstructing from candidates and (3) matching them with the sample.}
    \label{i:f2_architecture}
    \end{center}
    \vskip -0.2in
\end{figure*}

For the second research question, inspired by~\cite{marcel1983conscious}, the \emph{hypothesis-verification} process in human perception is simulated in the task of hand-written digit classification, where modules of an \textsc{estimator} and an \textsc{identifier} trained offline separately, are used either as auxiliaries of the basic \textsc{classifier} or as an independent architecture (Fig.~\ref{i:f2_architecture}).
It is shown in the result that, by leveraging the learned knowledge of mechanisms, the \textsc{estimator} and the \textsc{identifier} as auxiliaries can improve the classification accuracy significantly with extra explainability.
When the two modules operate independently of the \textsc{classifier}, without accessing any data of hand-written digits during training and through a pipeline of analyzing, reconstruction and matching, the architecture exceeds the performance of the basic \textsc{classifier}.

To our best knowledge, this is the first work that utilize the systematic knowledge about other mechanisms in classifying images. 
As a result, in addition to answer questions like ``Is there a `5' in the image? '', the proposed architecture is also able to answer ``Why do you think it is a `5'? '',  based on the knowledge it has been mastered, just like humans. 
The main contributions are as follows:
\begin{itemize}
    \item We demonstrate a learning methodology, with which the DNNs can learn the knowledge of specific mechanisms robustly using synthetic datasets constructed under the causal framework (and thus the first research question is answered).
    \item We design a novel architecture that simulates human visual perception in image classification, with additional explainability, based on the knowledge that has been mastered (and answers the second question).
\end{itemize}

\section{Related Work}

In this section, techniques and research topics in computer vision related to this work, are briefly reviewed.

\textbf{Data Augmentation and Domain Randomization.}
To tackle the potential drop in \textit{o.o.d}~performance, effective and commonly used techniques include data augmentation~\cite{goodfellow2016deep, shorten2019survey} and domain randomization~\cite{tobin2017domain, khirodkar2019domain, ren2019domain, amiranashvili2021pre}.
These two techniques share similar principles. 
The former technique is usually referred specifically to 2D transformations; the latter is adopted when manipulations are made on parameters in 3D environments. 
In a causal perspective, they both make treatment randomization to get rid of confounders and to improve the learning of invariance.
Based on this principle, this work also produces synthetic datasets through treatment randomization, but with a different purpose, that is, instead of randomizing \emph{out} the mechanisms, we aim to take them \emph{into} consideration in classification tasks.

\textbf{Parameter Estimation.}
As introduced previously, the task for learning mechanisms of 2D transformations is to estimate the transformation parameters. 
This task is extensively studied in various computer vision topics, such as 2D spatial invariance learning~\cite{jaderberg2015spatial}, text detection~\cite{xu2019lecture2note, xu2022morphtext, xu2022arbitrary, xu2024seeing}, and 3D pose estimations~\cite{madan2020capability, chen2020aware},  among many others.
However, in most of the existing studies, parameter estimation can only be performed restricted to object categories that appear in the training sets.
An important reason is that single-image parameter estimation is an ill-defined problem, in the sense that parameters in transformations are actually procedural variables, whose values are determined by both of the pre- and post-transformation states.
Therefore, models trained with methodologies based on single images, can hardly generalize to unseen categories.
In this work, the parameter estimation ability that we are interested in, should exhibit a certain degree of systematicity similar to human beings.
Another series of works~\cite{zhang2019aet, wang2019enaet} and the study in~\cite{locatello2020weakly} conduct representation learning based on pairs of images that are related through mechanisms, by using a single encoder for multiple mechanisms. 
However we try to isolate knowledge about single mechanisms and reuse them in downstream tasks.

\textbf{Program Induction.}
The knowledge learning in this work is essentially a program induction problem.
Active deep learning topics in this area include program synthesis~\cite{balog2016deepcoder, ellis2020dreamcoder}, image generation~\cite{lake2015human, young2019learning}, \textit{etc.} 
Program induction 
aims for more effective generation of programs, whereas 
this work focuses more on the interpretation of images that are beneficial for downstream tasks.
Therefore, the domain-specific languages in this work are fundamentally different, being more semantically relevant to the downstream tasks.

\section{Methodology}\label{s:methodology}

The aim of this work is to answer the two questions raised in the Introduction by investigating systematicity of the knowledge about the mechanisms and its application in hand-written digit classification.
During classification, the test set has a potential covariate shift caused by a target mechanism that is known but cannot be overcome through data augmentation techniques (which is a common situation in real-world tasks).
We simulate this setting by applying random 2D transformations on the MNIST~\cite{lecun1998gradient} test set, with no data augmentation operations of any kind performed during training. 

Inspired by the perception process in Fig.~\ref{i:f1_gestalt}, we propose that if machines learn the knowledge of a target mechanism, they could perform better in classification under covariate shift that is caused by the mechanism. 
Hence, an architecture is devised consisting of three DNN modules: a \textsc{classifier} $C$, an \textsc{estimator} $E$ and an \textsc{identifier} $D$, and thus abbreviated as ``CED''.
Causal datasets are constructed (in Section~\ref{s:datasets}) for the modules $E$ and $D$ to learn the knowledge of mechanisms (in Section~\ref{s:mechlearning}).
CED makes predictions in classification by raising hypotheses with $C$ and verifying them with $E$ and $D$.
The roles of these three modules are described in detail in Section~\ref{s:architecture}.

\subsection{Causal Datasets}\label{s:datasets}

\begin{figure}[t] 
    \vskip 0.15in
    \begin{center}
    \centerline{\includegraphics[width=0.6\columnwidth]{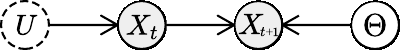}} 
    \caption[The causal graph of image transformation]{The causal graph of image transformation. $X_t$: Image at time step $t$ (before transformation). $X_{t+1}$: Image at time step $t+1$ (after transformation). $\Theta$: Parameter(s) of the transformation in study, as the variable is randomly sampled, this `treatment randomization' operation removes all arrows pointing to $\Theta$. $U$: Other unobservable variables that cause the generation of $X_t$.}
    \label{i:f3_causalgraph}
    \end{center}
    \vskip -0.2in
\end{figure}

To help DNNs learn the knowledge of a mechanism, the principle based on which a causal dataset is constructed, is explained below.
Denote by $\mathbf{x}_t$ and $\mathbf{x}_{t+1}$, respectively, the images before and after transformations $f$ (parameterized with $\theta$), \textit{i.e.}, 
\begin{equation}
    \mathbf{x}_{t+1} = f(\mathbf{x}_t ; \theta).
\end{equation}
As explained in the Introduction, the goal of knowledge learning is to estimate the value of $\theta$.
Let $X_t$, $X_{t+1}$ and $\Theta$ be the variables from which $\mathbf{x}_t$, $\mathbf{x}_{t+1}$ and $\theta$ are instantiated, respectively.
According to the causal graph in Fig.~\ref{i:f3_causalgraph}, if the estimation is made based only on the image \emph{after} transformation, \textit{i.e.}, $\mathbb{E}(\Theta|X_{t+1})$, given that $X_{t+1}$ is a collider, conditioning on it will inevitably cause the information flow from $U$ to $\Theta$, which will hinder us from learning robust knowledge of $f$ (via $\Theta$).
Therefore, in order to remove confounding caused by $U$, thus making the prediction of $\Theta$ more stable and generalize better in test domains, we have to condition on both $X_t$ and $X_{t+1}$, \textit{i.e.}, the Markov blanket of $\Theta$.
\footnote{This is also intuitively true, because it is pointless to ask how a picture has been transformed when no reference is provided.}

Concretely, in knowledge learning we aim to compute $\mathbb{E}_{P_{test}}(\Theta|X_t,X_{t+1})$ given only access to $P_{train}(\mathbf{x}_t,\mathbf{x}_{t+1},\theta)$, with the assumptions of: 
\begin{align*}
    &P_{train}(\theta|\mathbf{x}_t, \mathbf{x}_{t+1}) = P_{test}(\theta|\mathbf{x}_t, \mathbf{x}_{t+1}), \text{ and}\\
    &supp_{train}(\mathbf{x}_t, \mathbf{x}_{t+1}) = supp_{test}(\mathbf{x}_t, \mathbf{x}_{t+1}),
\end{align*}
where $P_{train}$ and $P_{test}$ are distributions of data in training and test domains, respectively, and $P_{train}(\mathbf{x}_t,\mathbf{x}_{t+1},\theta) \ne P_{test}(\mathbf{x}_t,\mathbf{x}_{t+1},\theta)$.

In this work, synthetic datasets for knowledge learning are constructed according to the above causal framework. 
Each data point is composed of a pair of images that are before and after transformations and the label $\theta$. Since the labels are automatically generated and no manual annotation is needed, this can be viewed as a self-supervised learning problem.

To acquire knowledge of mechanisms that is useful in classification with CED, one of the target mechanisms is the 2D transformation $f_T$ that includes rotation, scaling and translation, and data points are generated through $\mathbf{x}_{t+1} = f_T(\mathbf{x}_t; \theta_T)$ with affine transformation functions as $f_T$.
Another one is an identity function $f_I(\mathbf{x}_t; \theta_I)$ defined as:
\begin{equation}\label{e:identity}
    \mathbf{x}_{t+1} =
        \begin{cases}
            f_T(\mathbf{x}_t; \hat{\theta_T})   & \text{if } \theta_I = 1;\\
            f_T(\mathbf{x}'_t; \hat{\theta_T})  & \text{if } \theta_I = 0,
        \end{cases}
\end{equation}
where $\mathbf{x}'_t$ is a random sample other than $\mathbf{x}_t$, and $\hat{\theta_T}$ is the output of the module $E$ in CED (which will be explained in detail in Sections~\ref{s:mechlearning} and \ref{s:architecture}).

\subsection{Knowledge Learning}\label{s:mechlearning}

Based on the above causal datasets, the \textsc{estimator} $E$ and the \textsc{identifier} $D$ are trained to learn knowledge of 2D transformations $f_T$ and the identity function $f_I$, respectively.
Specifically, we employ $E$ that takes as the input paired images $\mathbf{x}^E_t$ and $\mathbf{x}^E_{t+1}$ generated from $f_T$ to predict the parameters $\hat{\theta_T}$.
The role of $D$, on the other hand, is to learn from $f_I$ and to predict the probability that two images are of the same identity.
In practice, the inputs of $D$ are $\mathbf{x}^D_t = \mathbf{x}^E_{t+1} = f_T(\mathbf{x}^E_t; \theta_T)$, and $\mathbf{x}^D_{t+1} = f_I(\mathbf{x}^E_t; \theta_I)$.

The mechanism of $f_T$ are independent of $f_I$, and thus $E$ is optimized first, by minimizing the mean squared error ($L_{MSE}$).
$D$ is then trained based on datasets generated with $f_I$ and $E$, and optimized by minimizing the binary cross entropy loss ($L_{BCE}$).
Therefore, the objectives of knowledge learning in this study can be represented as:
\begin{align}
    &\arg \min_E L_{MSE} (E(\mathbf{x}^E_t, f_T(\mathbf{x}^E_t; \theta_T)), \theta_T)\\
    &\arg \min_D L_{BCE} (D(f_T(\mathbf{x}^E_t; \theta_T), f_I(\mathbf{x}^E_t; \theta_I)), \theta_I)
\end{align}

\paragraph{Models} 

\begin{figure}[bt] 
    \vskip 0.1in
    \begin{center}
    \centerline{\includegraphics[width=0.7\columnwidth]{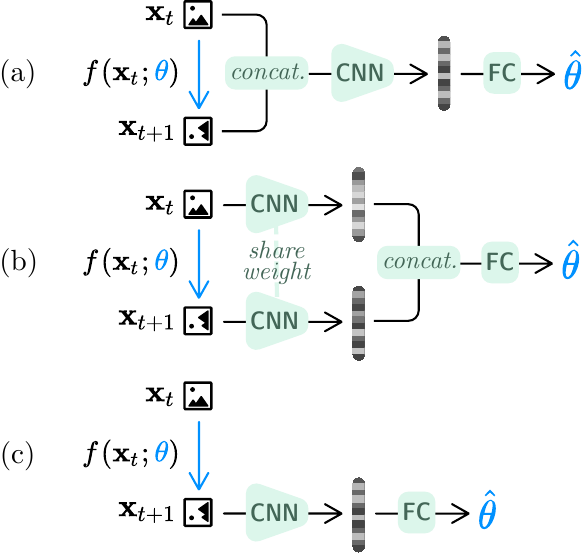}} 
    \caption{The models in knowledge learning. (\textbf{a}) FactorNet: $\mathbf{x}_t$ and $\mathbf{x}_{t+1}$ are concatenated in channel dimension before being fed into CNN; (\textbf{b}) Siamese Network: $\mathbf{x}_t$ and $\mathbf{x}_{t+1}$ are fed into CNN, whose outputs are then concatenated and sent to fully-connected (FC) layers; (\textbf{c}) Vanilla CNN: Only the transformed images $\mathbf{x}_{t+1}$ are fed into CNN.}
    \label{i:mnist_cnn}
    \end{center}
    \vskip -0.25in
\end{figure}

To obtain a more robust DNN in knowledge learning for modules $E$ and $D$, three Convolutional Neural Networks (CNNs) are investigated, as illustrated in Fig.~\ref{i:mnist_cnn}. 
The first model in study takes concatenated images as input (shown in Fig.~\ref{i:mnist_cnn}(a)).
It is called FactorNet for brevity in this work.
The following two models are relevant to this work, and thus are used as baselines.
The Siamese Networks~\cite{chopra2005learning} (shown in Fig.~\ref{i:mnist_cnn}(b)) are extensively studied on datasets with intrinsic relations in metric learning and representation learning.
Vanilla CNN (shown in Fig.~\ref{i:mnist_cnn}(c)) that takes single-images as input to make predictions, is another common method for numerical regression tasks. 

\subsection{Architecture CED for Classification}\label{s:architecture}

We now describe in detail each module in the proposed architecture CED for classification and their roles in simulating the ``hypothesis-verification'' process.

\textbf{Classifier $C$}.
The images in the MNIST test set are transformed before testing, denoted by $X^{test}_t$, while those in the training set are original ones without any transformation, denoted by $X^{train}_0$.
Given a test sample $\mathbf{x}^{test}_t \in X^{test}_t$, module $C$ produces a probability distribution across all classes, which is exploited as confidence scores. 
If the highest confidence score is lower than a preset threshold, instead of making a prediction of label, $C$ will output a hypothesis $H(\mathbf{x}^{test}_t) = \{y_i\}^k_{i=1}$, containing a list of class labels with top $k (k > 1)$ confidence scores. 

\textbf{Estimator $E$}.
Module $E$ randomly samples $N (N \geqslant 1)$ candidates from $X^{train}_0$ for each class in $H(\mathbf{x}^{test}_t)$.
Concretely, if the set of all candidates for $\mathbf{x}^{test}_t $ is denoted by $X_c \subset X^{train}_0$, we have $X_c = \{X^{(y_i)}_c\}^k_{i=1}$, and each $X^{(y_i)}_c = \{\mathbf{x}^{(y_i)}_j\}^N_{j=1}$.
With the assumption that $\mathbf{x}^{test}_t$ may be transformed from what looks similar to some of the candidates in $X_c$, $E$ then analyzes the relationship between $\mathbf{x}^{test}_t$ and each candidate \textit{w.r.t.} 2D transformation using knowledge learned previously, by computing $\widehat{\theta^{i,j}_T} = E(\mathbf{x}^{(y_i)}_j, \mathbf{x}^{test}_t)$.

\textbf{Identifier $D$}.
Since $E$ is a deterministic function and will produce an output, regardless of whether two images are really related, the role of $D$ is to examine which candidate is more closely related to the $\mathbf{x}^{test}_t$.
To achieve this, firstly, $D$ performs reconstructions on each candidate by exploiting the instructions from $E$ and obtains $ \mathbf{x}^{(y_i)}_{j,t} = f_T(\mathbf{x}^{(y_i)}_j; \widehat{\theta^{i,j}_T})$.
Then, $\mathbf{x}^{(y_i)}_{j,t}$ is tested on how likely it matches to $\mathbf{x}^{test}_t$, by leveraging the knowledge learned about the identity function $f_I$.
The label of the candidate with highest likelihood will be output as the final prediction $\hat{y} = \arg \max_{y_i} D(\mathbf{x}^{test}_t, \mathbf{x}^{(y_i)}_{j,t})$.

In the above process, potential classes are hypothesized by $C$, and verification on these classes is made by modules $E$ and $D$ through the pipeline of (a) analyzing possible transformations, (b) reconstructing from candidates and (c) matching them with the sample.

\textbf{ED}.
It can also be noticed that the pre-trained modules $E$ and $D$ do not have to access MNIST during training, and do not rely on $C$ too much either.
Based on the fact that the training and test set of MNIST share the same class label space, we also explore a second architecture that only employs $E$ and $D$ (abbreviated as ``ED'').
The only difference from CED is that ED directly takes all classes as hypothesis ($k = 10$).

\section{Experiments}\label{s:experiments}

In this section, experiments are conducted to answer the two questions raised in the Introduction. 

\subsection{Is the Learned Knowledge Systematic?}

In order to study the robustness of estimation on $\theta_T$ and $\theta_I$ of $f_T$ and $f_I$, synthetic datasets are constructed according to Section~\ref{s:datasets}.
Three DNN models are trained and tested based on the methodology illustrated in Section~\ref{s:mechlearning}.
We now describe specifically how the experiments are conducted.

\subsubsection{Learning of 2D Image Transformation mechanisms}\label{s:learn_trans}

\textbf{Datasets.}
In the experiments, the original images in MNIST, EMNIST~\cite{cohen2017emnist} and  CIFAR-10~\cite{krizhevsky2009learning} are used as $\mathbf{x}_0$.
To eliminate potential overfitting, we obtain the input image pairs $\mathbf{x}_t = f_T(\mathbf{x}_0; \theta^0_T)$ and $\mathbf{x}_{t+1} = f_T(\mathbf{x}_t; \theta_T)$, where $\theta^0_T$ and $\theta_T$ are randomly sampled in a uniform distribution (see Table~\ref{t:range_parameters}).

In this work, we conduct learning on four types of $f_T$, including the individual learning of rotation, scaling and translation, and the joint learning of all the above three.
For individual learning, only one of the three transformations is applied on $\mathbf{x}_t$ at a time, while all three transformations are applied simultaneously for joint learning.

To further increase the difficulty of the task, a synthetic dataset composed of black/white noises (of a Bernoulli distribution) is randomly generated and used as $\mathbf{x}_t$.
To better test robustness, all test data are sampled from datasets that are semantically different from the training sets. 
The detailed schemes are listed in Table~\ref{t:datasets}.

\begin{table}[tb]
\caption{The parameters of 2D transformations. The values of each parameter are uniformly sampled within their ranges.}
\label{t:range_parameters}
\vskip 0.1in
\begin{center}
\begin{small}
\begin{tabular}{ll}
\hline
Parameter                & Range \\
\hline
Rotation angle           & $[-90^{\circ}, 90^{\circ}]$ \\
Translation (horizontal) & $[-5, 5]$ pixels\\
Translation (vertical)   & $[-5, 5]$ pixels\\
Scale factor             & $[0.7, 1.3]$\\
\hline
\end{tabular}
\end{small}
\end{center}
\vskip -0.15in
\end{table}

\begin{table}[tb]
\caption{The training and test data used for knowledge learning.}
\label{t:datasets}
\vskip 0.1in
\begin{center}
\begin{small}
\begin{tabular}{lll}
\hline
Experiment  &       & Dataset \\
\hline
Exp\_MNIST  & Train & training set in MNIST\\
            & Test  & `letter' division of test\\
                    &&set in EMNIST\\
\hline
Exp\_CIFAR  & Train & 9 classes of training set\\
                    &&in CIFAR-10\\
            & Test  & the other class of\\
                    &&training set in CIFAR-10\\
\hline
Exp\_NOISE  & Train & black/white noise\\
            & Test  & test set in MNIST\\
\hline
\end{tabular}
\end{small}
\end{center}
\vskip -0.15in
\end{table}

\textbf{Models.}
The CNN backbone in~\cite{zhang2019aet} is used in all three models in Fig.~\ref{i:mnist_cnn}.
All input pairs of $\mathbf{x}_t$ and $\mathbf{x}_{t+1}$ are concatenated along the channel dimension before being fed in to the FactorNets; the input size is $N_{batch}\times2\times28\times28$ in Exp\_MNIST and Exp\_NOISE, and $N_{batch}\times6\times32\times32$ in Exp\_CIFAR, where $N_{batch}$ is the batch size.

The performance of FactorNets for 2D transformations learning is reported in Fig.~\ref{i:perform_mechnet_ro} and Fig.~\ref{i:perform_mechnet_sc_tr} in Appendix~\ref{a:2D}. 
For the learning of individual mechanisms, it can be observed that most of the absolute percentage errors (APE) (\textit{e.g.} the third quartile in the distributions) can be controlled to below $20\%$ in most of the experiments, and even $10\%$ in Exp\_CIFAR.
The accurate parameter estimation suggests effective learning of 2D transformation knowledge.
Furthermore, there are only minor differences between the distributions of APE when we compare the training and test sets.
This can be attributed to the strong \textit{o.o.d.}~systematicity of knowledge learned with FactorNets, given that the data in the training and test sets are completely different in semantics.
More results on the performance of FactorNets for 2D transformation learning are shown in Appendix~\ref{a:2D}.

\begin{figure}[bt] 
    \vskip 0.1in
    \begin{center}
    \centerline{
    \includegraphics[width=0.72\columnwidth]{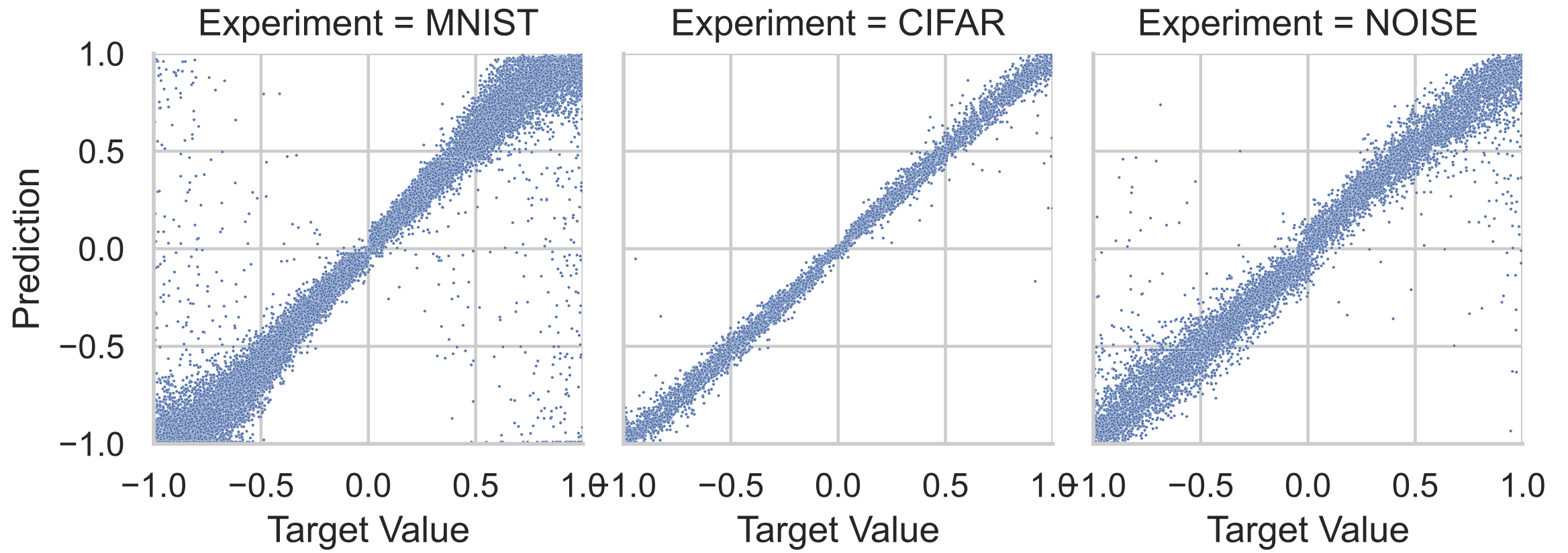}
    \includegraphics[width=0.28\columnwidth]{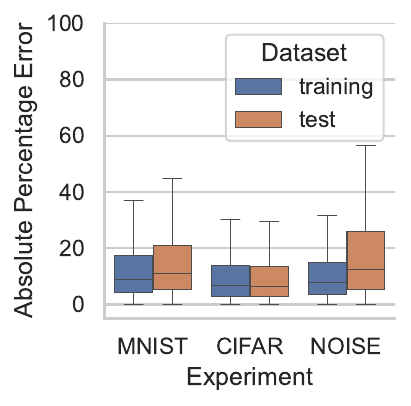}}
    \caption{Performance of FactorNets for individual rotation learning. (\textbf{left}) Predictions of rotation angle \textit{vs.}~the ground truth (normalized to $[-1,1]$) in test set. (\textbf{right}) Distributions of absolute percentage errors (in \%) of all data points in the dataset. }
    \label{i:perform_mechnet_ro}
    \end{center}
    \vskip -0.2in
\end{figure}

\subsubsection{Learning of the Identity Function}

To evaluate the \textit{o.o.d.}~performance of the $\theta_I$ estimation, we train FactorNet in Exp\_NOISE using samples generated with the method in Section~\ref{s:datasets}, and $\theta_I$ is randomly sampled in $\{0,1\}$ in order to produce a balanced dataset. 
The FactorNet for rotation learning trained in Exp\_NOISE is used as $E$.
The resulting F1 scores are $0.9987$ and $0.9757$ for training and test set, respectively, which indicates superior \textit{o.o.d.}~performance of FactorNet for identification tasks as the module $D$ in CED.

\subsubsection{Key Elements in Knowledge Learning}\label{s:key_elements}

In this section, several ablation studies are conducted to examine elements 
crucial for a robust knowledge learning.

\begin{figure}[bt] 
    \vskip 0.1in
    \begin{center}
    \centerline{
    \includegraphics[width=\columnwidth]{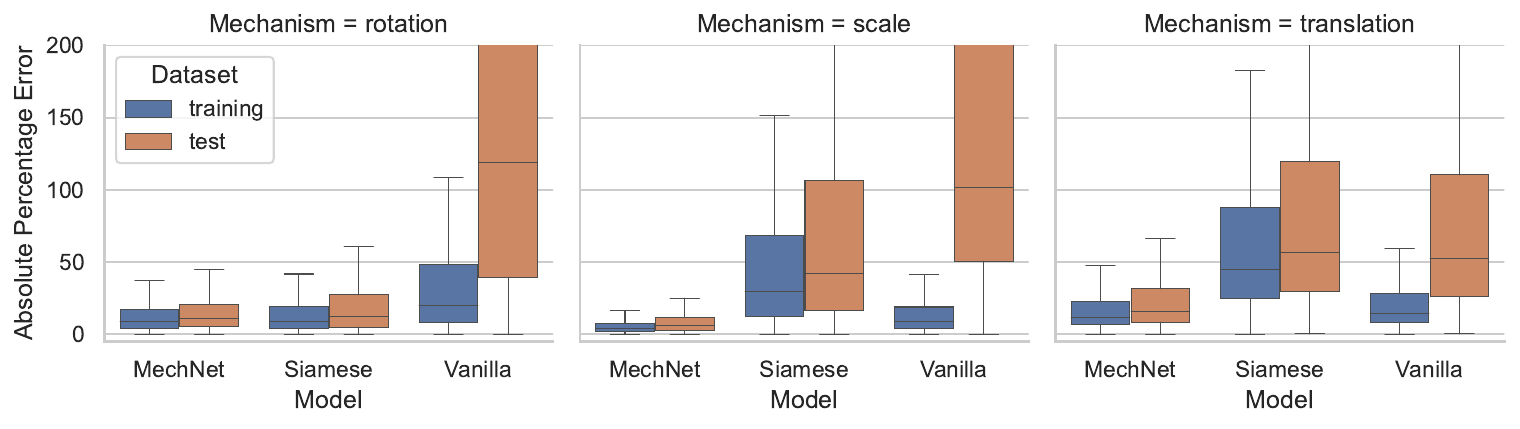}
    } 
    \caption{The performance of individual transformation learning across different models.}
    \label{i:perform_models_tran}
    \end{center}
    \vskip -0.2in
\end{figure}

Firstly, as analyzed based on the causal graph in Fig.~\ref{i:f3_causalgraph}, if there exists causal relationship from $U$ to $X_t$, it is necessary to condition on both $X_t$ and $X_{t+1}$ in order to predict $\Theta$ robustly. 
As shown in Fig.~\ref{i:perform_models_tran}, the \textit{o.o.d.}~performance gap of vanilla CNN is noticeable in all learning cases, compared with FactorNet and Siamese networks that take both $X_t$ and $X_{t+1}$ as inputs.
Vanilla CNN performs relatively better in translation learning, because the position of $X_t$ (the original images in this case) is always in the center and independent of $U$. 
However, while being able to estimate rotation angles accurately in the training set, vanilla CNN completely fails in the test set, because the estimation of angles relies highly on the pattern of images, which is determined by $U$.
This also offers insight into numerical regression tasks in contemporary computer vision studies, such as object pose estimation; given only the images after transformation during training, a good \textit{o.o.d.}~performance cannot be expected.

Secondly, for CNN backbones, computations based on concatenated images are necessary to make more accurate estimations.
Fig.~\ref{i:perform_models_tran} shows that Siamese networks underperform FactorNets in all mechanisms.
Much information about transformations is lost through convolutional operations and the max pooling layers, while more information can be preserved from the beginning in FactorNets.

Additionally, we speculate that the inductive bias of CNNs fundamentally affects the effectiveness of knowledge learning.
This is based on the observation of the learning curves of the three mechanisms (in Fig.~\ref{i:learning_curve_models_tran} in Appendix~\ref{a:IB}).
Fast learning on translation and scaling and a slow one on rotation can be noticed for all models, which indicates that CNN models have greater difficulty learning the mechanism of rotation.
Considering CNN properties of translation-equivariant, positional information can be encoded and operated with CNN at higher efficiency. 
An extensive investigation into other inductive bias is necessary for a more solid claim to be made in the future.

\subsection{Can Knowledge be Leveraged?}

In the previous section, it can be seen that effective learning can be achieved with FactorNets.
The models are capable of making accurate estimations on parameters $\theta$, and this capability can be generalized to semantically different datasets. 
This indicates a certain level of systematicity.
Hence, with these models as building blocks, we construct the CED architecture according to Section~\ref{s:architecture}. 
Comparison of classification performance amongst different architectures are conducted first (in Section~\ref{s:class_performance}), followed by discussion of the simulation of human-like visual perception (in Section~\ref{s:simulation}). 

\subsubsection{Classification Performance}\label{s:class_performance}

In the experiment, classification is performed with the setting of covariate shift caused by rotation, and the performance is compared amongst a basic classifier, CED and ED.
To construct CED and ED, FactorNets trained in Exp\_NOISE for the (individual) rotation learning and the identity function learning are exploited as the module $E$ and $D$, respectively.
The basic classifier $C$ is trained with original images $X^{train}_0$ in MNIST without any data augmentations.
The length of hypothesis $H(\mathbf{x}^{test}_t)$ is $k=5,10$ for CED and $k=10$ for ED.
The number of candidates for the $E$ is $N=200$ for each class.
The confidence threshold of $C$ is set to $0.9999$.  

\begin{figure}[bt] 
    \vskip 0.1in
    \begin{center}
    \centerline{
    \includegraphics[width=0.85\columnwidth]{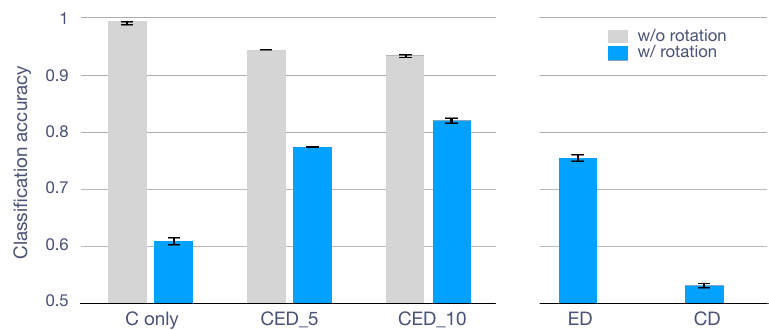}
    } 
    \caption{The performance of classification. CED\_5 and CED\_10 denote CED with hypothesis $k=5$ and $k=10$, respectively.}
    \label{i:perform_class}
    \end{center}
    \vskip -0.2in
\end{figure}

The classification accuracy obtained on the MNIST test set, with or without rotations, is shown in Fig.~\ref{i:perform_class}.
The first observation is that, in the case of rotated test set, the basic classifier has experienced nearly a $40\%$ performance drop.
However, the accuracy of CED has increased to $77\%$ when $k=5$ (CED\_5) and further to $82\%$ when $k=10$ (CED\_10).
In CED, $E$ and $D$ are introduced for further interpretation when $C$ is not very confident, and they provide extra explanations about why the sample is classified as such and how it is rotated, by leveraging the knowledge about rotation with $E$. 
Additionally, this process does not affect the performance too much for the test set without rotation.

ED outperforms the basic classifier by classifying with an accuracy of more than $75\%$.
It is worth noting that the performance is achieved without any knowledge of the handwritten digits (since both $E$ and $D$ are trained in Exp\_NOISE), but only through the processes of analyzing, reconstructing and matching.
Furthermore, only $4\%$ ($200 \times 10 / 50000$) of the training data are accessed during inference.
This is behaviorally similar to human beings, who are capable of classifying characters that they do not know at all, so long as necessary references are provided.

To investigate further the role of $E$ with its knowledge about rotation, an ablation study was conducted on CD by removing $E$ from the CED.
Obviously, the CD architecture loses the ability to interpret transformation and the performance on rotated test set has dropped to below $60\%$ (Fig.~\ref{i:perform_class}).
On one hand, this indicates the importance of rotation knowledge to $D$, which requires the instructions for reconstruction;
On the other hand, since the rotated samples look very different from the candidates, indirectly it also demonstrates the effectiveness of $D$.

\begin{figure}[bt] 
    \vskip 0.1in
    \begin{center}
    \centerline{
    \includegraphics[width=0.85\columnwidth]{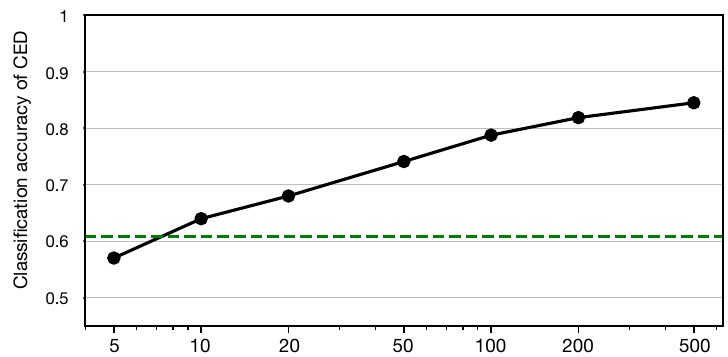}
    } 
    \caption{The classification accuracy of CED with different numbers of candidates. Performance already surpasses the basic classifier (the green dash line) when $N \geqslant 10$.}
    \label{i:perform_num_candidates}
    \end{center}
    \vskip -0.2in
\end{figure}

\textbf{The number of candidates.}
As shown in Fig.~\ref{i:perform_num_candidates}, classification accuracy is greatly affected by the number of candidates. 
Given that $D$ is trained on noise, the module is really sensitive to nuance differences. 
Therefore, in order to find a candidate that is very similar to a sample, a very large candidate pool is required.

In addition, the generation of digits can also be viewed as a mechanism. 
Unlike 2D transformations, the parameterization of digit generation is much more complicated~\cite{lake2015human}. 
While the integration of an estimation module for digit generation (as a new $E$) into the existing CED would presumably reduce the required number of candidates significantly, this will, at the same time, introduce new challenges in compositionality, which involves the collaboration between multiple $E$s.

\subsubsection{Simulation of Human-like Visual Perception}\label{s:simulation}

In this work, we propose CED as a preliminary simulation of the ``hypothesis-verification'' process~\cite{marcel1983conscious} in human visual perception.
Although the simulation is not a reverse engineering of the human brain, based on psychological studies about cognition and behaviors, both human and CED share similarities in how information is processed.

As human beings, we have powerful ability to model an object with functionally easier mechanisms according to Gestalt principles~\cite{koffka2013principles}.
This does not happen only in visual perception, but also in other aspects of behaviors~\cite{Gazzaniga1998split, nisbett1977telling}, where people try to rationalize their behaviors with convincing (but sometimes incorrect) reasons.
The role of $E$ and $D$ in CED is actually to provide explainability, with which machines can ``make sense'', to some extent, of what they see.
This explainability also provides possibilities for humans to improve the architectures, in ways that they can comprehend.

Furthermore, the simulation and imagination in the brain have been studied in various works, and are proposed as the key elements in the understanding of physical scenes and counterfactual reasoning~\cite{battaglia2013simulation, pearl2018book}. 
Based on the model of the world in the mind, humans can make predictions about the future (in causal direction) and infer the causes of things that have happened (in anti-causal direction). 
In the architecture of CED, simulations of 2D-transformations in anti-causal and causal directions are enabled with module $E$ and the affine transformation function, respectively, which equip the machine with an imagination space.




\section{Conclusion and Future Work}
To conclude, in this study, stable knowledge learning has been shown to be possible if models have been trained on (concatenated) data pairs that are intrinsically related through the mechanism. 
Based on this learning methodology, FactorNets with their acquired knowledge play significant roles in image classifications and further interpretations under covariate shift.
The performance boost of the proposed CED architecture also suggests the effectiveness of the simulation to human-like visual perception.
We hope the simulation, along with its basis, \textit{i.e.}, the learning methodology, can provide inspirations for future studies in computer vision \textit{w.r.t.}~human-like general AI.

Based on the findings in this work, we identify some limitations and questions that are appealing for further investigations. 

\textbf{Compositionality:} In this work, covariate shift is introduced in test set by intervening on only one mechanism (\textit{i.e.}~rotation).
In the setting where multiple mechanisms are considered, it will be ideal if multiple $E$s could leverage the knowledge learned separately and cooperate with each other. 
However, some preliminary results show that $E$s will not generalize well, if the training is based on only the interventions of target mechanism and keeping the others fixed.
This is in line with~\cite{madan2020capability}, where the generalization improves only if more combinations of two mechanisms (category and pose) are exposed \emph{during training}.
Therefore, essential architectural elements that would facilitate the communications and interactions between modules (especially $E$s) are intriguing for us to explore in the future. 

\textbf{3D Virtual World:} If we think of the real-world photos as the result of the interactions of mechanisms, such as foreground and background objects, lighting conditions, camera attributes, \textit{etc.}, then tasks based on real photos could also be tackled in the same manner as in this work.
With the rapid development of computer graphics, photo-realistic synthetic datasets with 1) controlled interventions on target mechanisms and 2) automatic pixel-accurate annotations can be efficiently created with 3D rendering engines.
As described in Section~\ref{s:datasets}, if the mechanism is stable across both virtual and real worlds, the knowledge learned on synthetic images could presumably be usable on real photos.
More importantly, the real-time rendering capability of modern game engines (\textit{e.g.} Unreal Engine) offers potential realization of an imagination space for machines (analogous to the affine transformation functions in the work).


\bibliographystyle{plain}
\bibliography{ref.bib}

\newpage
\appendix
\section{Additional Results}

\begin{figure}[hbt] 
    \vskip 0.1in
    \begin{center}
    \centerline{\includegraphics[width=\textwidth]{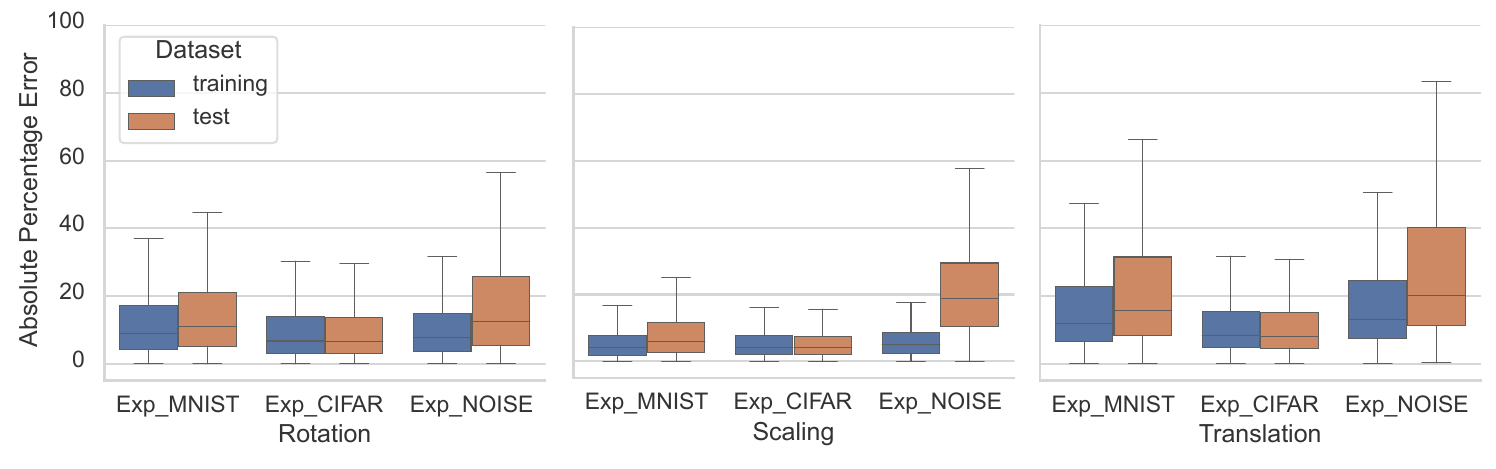}} 
    \caption{Performance of FactorNets for individual 2D transformation learning. \textbf{(left)} Rotation. \textbf{(center)} Scaling. \textbf{(right)} Translation.}
    \label{i:perform_mechnet_sc_tr}
    \end{center}
    \vskip -0.2in
\end{figure}

\paragraph{Individual learning.}\label{a:2D}
Additional results of performance of FactorNets for individual 2D transformation learning is shown in Fig.~\ref{i:perform_mechnet_sc_tr}.
Similar to the result in Fig.~\ref{i:perform_mechnet_ro}, several observations for individual learning are listed as follows.
\begin{itemize}
    \item Majority of absolute percentage errors can be controlled to below $20\%$ for individual learning, which indicates the effectiveness of 2D transformation learning.
    \item There are only minor differences in the distributions of absolute percentage error between the training and test sets for individual learning across all experiments, which suggests strong \textit{o.o.d.}~generalization.
\end{itemize}

\begin{figure}[hbt] 
    \vskip 0.1in
    \begin{center}
    \centerline{\includegraphics[width=\textwidth]{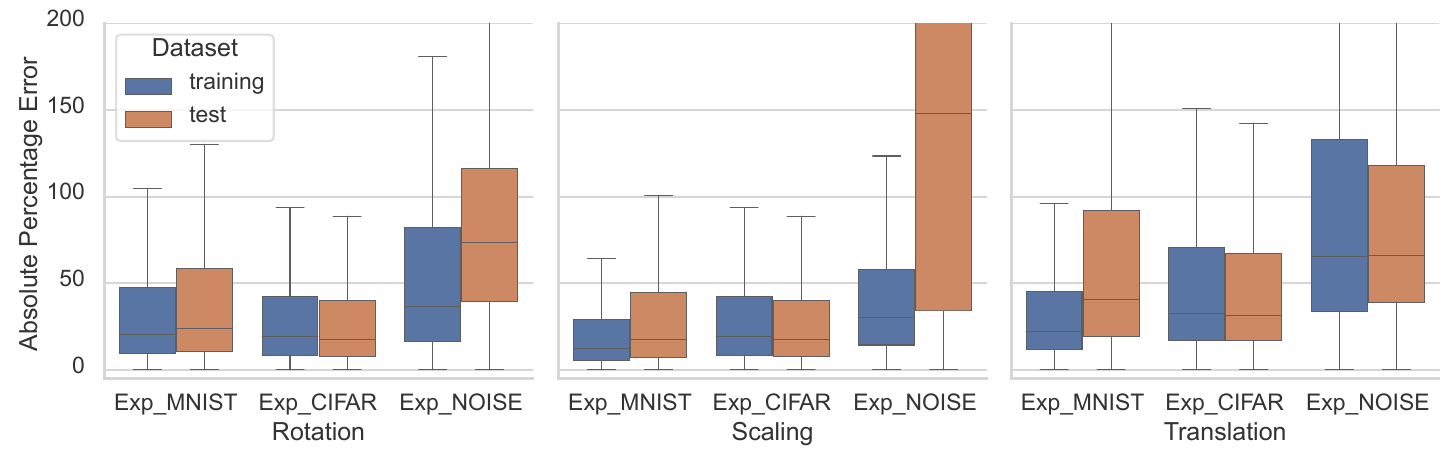}} 
    \caption{Performance of FactorNets for joint 2D transformation learning. \textbf{(left)} Rotation. \textbf{(center)} Scaling. \textbf{(right)} Translation.}
    \label{i:perform_mechnet_j}
    \end{center}
    \vskip -0.2in
\end{figure}

\paragraph{Joint learning.}
For joint learning of 2D transformation, obvious performance drop in both the training and test set can be observed in Fig.~\ref{i:perform_mechnet_j}, compared with the individual learning, even if the number of parameters of FactorNets is four times that of models for individual learning.
Similar results are reported in study~\cite{madan2020capability}, where more accurate estimations of variables are made by separately trained models, because of the improved ``selectivity and invariance at the individual neuronal level''.

\paragraph{FactorNets trained in Exp\_NOISE.} 
Although FactorNet exhibits strong \text{o.o.d.}~generalization, the performance decreases to some extent when the difference between the training and test sets becomes considerably big.
For instance, a larger performance gap between the training and test set in Exp\_NOISE can be noticed, compared with the other two experiments in Fig.~\ref{i:perform_mechnet_sc_tr} and \ref{i:perform_mechnet_j}.
The most apparent characteristic in this experiment is the pattern difference between noises and hand-written digits, which implies the potential difference in exploitation of patterns during learning.

\begin{figure}[hbt] 
    \vskip 0.1in
    \begin{center}
    \centerline{
    \includegraphics[width=0.7\textwidth]{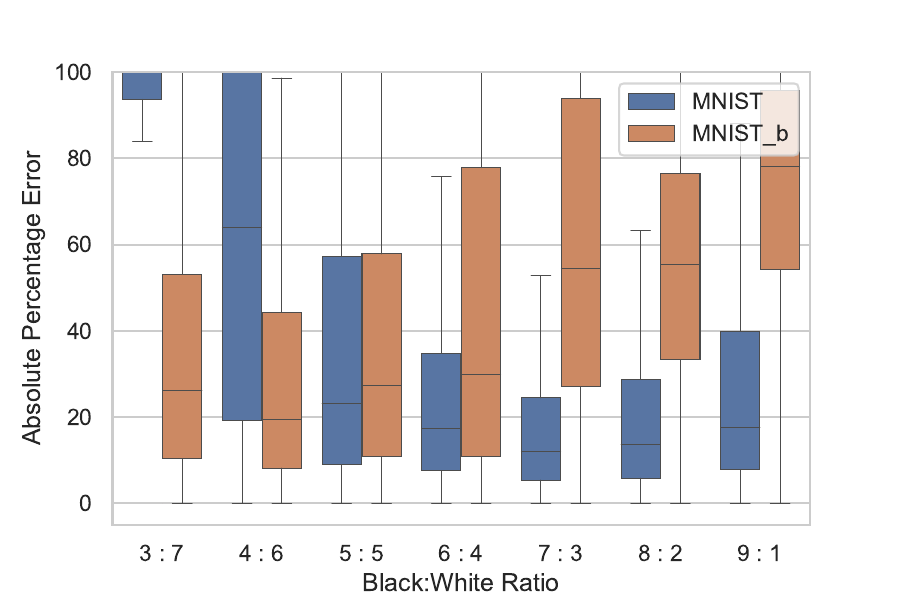}
    } 
    \caption{Performance of FactorNets in rotation learning with controlled black/white pixel ratios in EXP\_NOISE. Pixel values are swapped in MNIST\_b.}
    \label{i:perform_mechnet_bwratio}
    \end{center}
    \vskip -0.2in
\end{figure}

To prove this, an ablation study was conducted by altering the ratio of black to white pixels of the training data in Exp\_NOISE.
As shown in Fig.~\ref{i:perform_mechnet_bwratio}, the best-performing model for rotation learning is trained on $7:3$ black/white noises.
However, if the pixel values in MNIST are swapped (~\textit{i.e.}~black digits on white background), the best performance can be achieved around $4:6$.
Different ratios will provide different patterns that can be exploited in learning. 
The best ratio for individual learning of translation and rotation is around $7:3$, while for scaling it is around $3:7$, which can also explain the poor \textit{o.o.d.}~generalization performance of joint learning in Exp\_NOISE, since it is impossible for the model to learn the three transformations equally well with only one ratio.

\begin{figure}[hbt] 
    \vskip 0.1in
    \begin{center}
    \centerline{
    \includegraphics[width=0.85\textwidth]{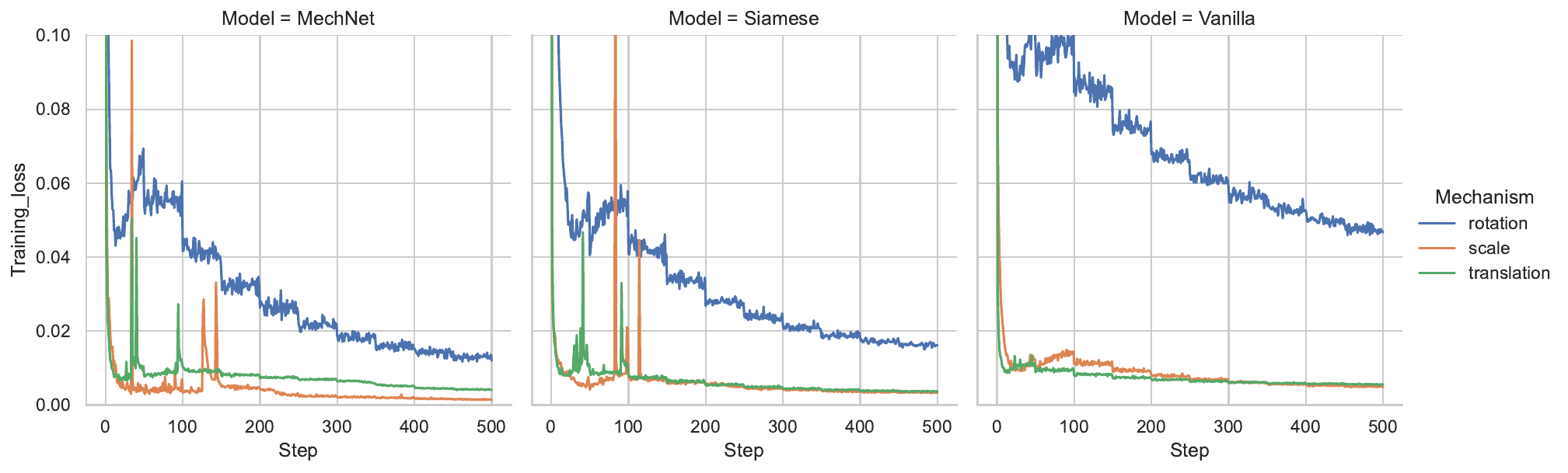}
    } 
    \caption{The learning curves in transformation learning across different models. Fast learning on translation and scaling and a slow one on rotation can be observed for all models.}
    \label{i:learning_curve_models_tran}
    \end{center}
    \vskip -0.2in
\end{figure}

\paragraph{Learning curves in 2D transformation learning.}\label{a:IB}
The learning curves in 2D transformation learning are shown in Fig.~\ref{i:learning_curve_models_tran}. 
For all three models, fast learning on translation and scaling and a slow one on rotation can be observed.

\begin{figure}[bt] 
    \vskip 0.1in
    \begin{center}
    \centerline{
    \includegraphics[width=0.5\columnwidth]{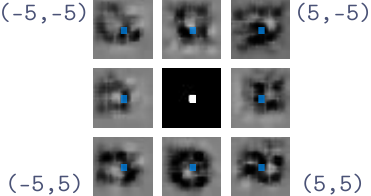}
    } 
    \caption{Images obtained with the Translation FactorNet through gradient decent. The image in the center is the original one $\mathbf{x}_t$. According to the values of $\theta$ (four of them are marked on the corners), $\mathbf{x}_{t+1}$ are generated through gradient decent. In each of $\mathbf{x}_{t+1}$, an obvious offset of the light area from the original position (the blue dot) to the target position can be observed.}
    \label{i:gradient_decent}
    \end{center}
    \vskip -0.2in
\end{figure}

\paragraph{Restoration.}
An interesting property of FactorNet and Siamese networks can be found further in translation learning.
Given an image $\mathbf{x}_t$ with a small square in the center, an image $\mathbf{x}_{t+1}$ identical to $\mathbf{x}_t$ and the target value of translation $\theta$, we can obtain a (coarse) translated version of $\mathbf{x}_t$ by optimizing $\mathbf{x}_{t+1}$ with gradient decent according to:
\begin{equation}\label{e:gradient}
    \mathbf{x}_{t+1} \leftarrow \mathbf{x}_{t+1} - \alpha \nabla_{\mathbf{x}_{t+1}}L_{MSE} (E(\mathbf{x}_t, \mathbf{x}_{t+1}), \theta),
\end{equation}
where $\alpha$ is the learning rate.
As shown in Fig.~\ref{i:gradient_decent}, this operation can be viewed as an approximation of the translation function $f_T$.
Although this reversed generation of images is by no means accurate and only limited to very simple patterns, the phenomenon cannot be repeated in cases of rotation and scaling. 

\section{Model Architecture Details}
We follow the implementation in~\cite{zhang2019aet} to construct the three models (in Fig~\ref{i:mnist_cnn}) for knowledge learning experiments.
The architectures for individual mechanism learning are shown in Table~\ref{t:models}.
The models for joint learning are only different in channel sizes, which are all doubled in Exp\_MNIST and Exp\_NOISE, and 50\% larger in Exp\_CIFAR.

\begin{table}[htb]
\caption{Architecture of models for knowledge learning.}
\label{t:models}
\vskip 0.15in
\begin{center}
\begin{small}
\begin{tabular}{ll}
\hline
Models in Exp\_MNIST and Exp\_NOISE  &  Models in Exp\_CIFAR \\
\hline
5$\times$5 Conv 96, BatchNorm, ReLU & 5$\times$5 Conv 192, BatchNorm, ReLU \\
1$\times$1 Conv 64, BatchNorm, ReLU & 1$\times$1 Conv 128, BatchNorm, ReLU \\
1$\times$1 Conv 32, BatchNorm, ReLU & 1$\times$1 Conv 64, BatchNorm, ReLU \\
3$\times$3 MaxPooling stride 2 & 3$\times$3 MaxPooling stride 2\\
3$\times$3 Conv 32, BatchNorm, ReLU & 3$\times$3 Conv 128, BatchNorm, ReLU \\
1$\times$1 Conv 32, BatchNorm, ReLU & 1$\times$1 Conv 128, BatchNorm, ReLU \\
1$\times$1 Conv 32, BatchNorm, ReLU & 1$\times$1 Conv 128, BatchNorm, ReLU \\
3$\times$3 MaxPooling stride 2 & 3$\times$3 MaxPooling stride 2\\
3$\times$3 Conv 32, BatchNorm, ReLU & 3$\times$3 Conv 128, BatchNorm, ReLU \\
1$\times$1 Conv 32, BatchNorm, ReLU & 1$\times$1 Conv 128, BatchNorm, ReLU \\
1$\times$1 Conv 32, BatchNorm, ReLU & 1$\times$1 Conv 128, BatchNorm, ReLU \\
3$\times$3 MaxPooling stride 2 & 3$\times$3 MaxPooling stride 2\\
2$\times$2 Conv 32, BatchNorm, ReLU & 2$\times$2 Conv 128, BatchNorm, ReLU \\
1$\times$1 Conv 32, BatchNorm, ReLU & 1$\times$1 Conv 128, BatchNorm, ReLU \\
1$\times$1 Conv 32, BatchNorm, ReLU & 1$\times$1 Conv 128, BatchNorm, ReLU \\
3$\times$3 MaxPooling stride 2 & 3$\times$3 MaxPooling stride 2\\
FC & FC\\
FC (Siamese Networks only) & FC (Siamese Networks only)\\
\hline
\end{tabular}
\end{small}
\end{center}
\vskip -0.1in
\end{table}

\end{document}